\documentclass[conference]{IEEEtran}
\IEEEoverridecommandlockouts
\usepackage{cite}
\usepackage{amsmath,amssymb,amsfonts}
\usepackage{algorithmic}
\usepackage{graphicx}
\usepackage{textcomp}
\usepackage{xcolor}
\usepackage[numbers,sort&compress]{natbib}
\def\BibTeX{{\rm B\kern-.05em{\sc i\kern-.025em b}\kern-.08em
    T\kern-.1667em\lower.7ex\hbox{E}\kern-.125emX}}
    
\begin{document}

\title{Aggregating Low Rank Adapters in Federated Fine-tuning\\

\thanks{Many thanks to the Apheris AI company for supporting this work and providing us with the computational setup for the conducted experiments.}
}

\author{\IEEEauthorblockN{Evelyn Trautmann}
\IEEEauthorblockA{\textit{Apheris AI} \\
Berlin, Germany \\
e.trautmann@apheris.com}
\and
\IEEEauthorblockN{Ian Hales}
\IEEEauthorblockA{\textit{Apheris AI} \\
Berlin, Germany}
\and

\IEEEauthorblockN{Martin F. Volk}
\IEEEauthorblockA{\textit{Apheris AI} \\
Berlin, Germany}
}

\maketitle

\begin{abstract}
Fine-tuning large language models requires high computational and memory resources, and is therefore associated with significant costs. 
When training on federated datasets, an increased communication effort is also needed. For this reason, parameter-efficient methods (PEFT) are becoming increasingly important. In this context, very good results have already been achieved by fine-tuning with low-rank adaptation methods (LoRA). The application of LoRA methods in Federated Learning, and especially the aggregation of adaptation matrices, is a current research field.
In this article, we propose a novel aggregation method and compare it  with different existing aggregation methods of low rank adapters trained in a federated fine-tuning of large machine learning models and evaluate their performance with respect to selected GLUE benchmark datasets.
\end{abstract}


\section{Introduction}
With the increasing importance of large language models (LLMs), even fine-tuning tasks have become more and more challenging. 
Computational cost has not only financial but also environmental and legal impact. For instance the EU AI Act clearly states energy sustainability as a requirement.
To save machine capacity and cost, parameter efficient fine-tuning (PEFT) has gained popularity, especially as results from PEFT methods have been shown to be reasonably performant keeping a high model quality while at the same time they achieve a considerable reduction of training parameters \cite{han2024parameterefficientfinetuninglargemodels}.

Low Rank Adaptation (LoRA \cite{hu2021lora}) is a popular PEFT method that aims to train only a low-dimensional approximation of the gradient in each fine-tuning iteration instead of the weight matrices. For this purpose, the actual weight matrices are frozen, and only two low-rank adapter matrices are trained, whose product approximates the gradient.

If the training data is not present at a single central place but distributed over several clients, it can still be used for fine-tuning. Federated Learning \cite{mcmahan2023communicationefficient} is a machine learning approach where a central server coordinates training across multiple decentralized devices or servers holding local datasets, without exchanging the data itself. This method enhances data privacy and security, as raw data remains on local devices, while only model updates are shared and aggregated. This setup alone does not guarantee privacy, but it allows for employment of privacy enhancement technologies as stated in subsection \ref{pet}.

PEFT methods like LoRA can be combined with Federated Learning. In an immediate approach all trained parameters are aggregated with Federated Averaging. We will show in this article the shortcomings of the naive approach and present an improved aggregation method. 

\subsection{Outline}
In the following, we will outline our method and conduct various experiments, in which we apply LoRA in a federated way where the data is distributed over several clients. 
The trained parameters must be aggregated at regular intervals. 
We show empirical evidence that our approach provides improved convergence in the early stages of model training, where the contribution of the A matrix is stronger \cite{pmlr-v235-hao24a, pmlr-v235-hayou24a}.
Finally, we discuss possible combinations of the individual methods.

\section{Background}

\subsection{Low Rank Decomposition}

Fine-tuning LLMs comes with several challenges. As models grow, the number of training parameters increases, leading to high training costs and deployment challenges. 

Against this backdrop, a dedicated research area focusing on parameter-efficient fine-tuning has emerged, encompassing methods such as adapter layers and prompt optimization. The LoRA method proposed in \cite{hu2021lora} represents a good trade-off between efficiency and model quality.

\begin{figure}
        \centering
        \includegraphics[width=5cm]{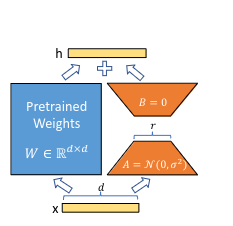}
        \caption{Low rank adaptors approximation weight increments, illustration
        from original paper \cite{hu2021lora}.}
        \label{fig:lora}
    \end{figure}

Whereas previous PEFT methods \cite{houlsby2019parameterefficient, pfeiffer-etal-2020-adapterhub}, merely focused on adding adapter layers to the model, LoRA modifies existing layers using low-rank approximations of weight increments without adding new layers. Each frozen layer is modified as illustrated in Fig. \ref{fig:lora}.

At its core, this method involves training two low-rank adapters, \(A\) and \(B\), of size \((m \times r)\) and \((r \times n)\), instead of the full \((m \times n)\) parameter matrix \(W\) for each layer. The product of these adapters approximates the gradient of the parameter matrix. This reduces the number of parameters to be trained from \(m {\cdot} n\) to \((m+n) \cdot r\), which leads to a significant reduction in the number of trainable parameters when \(r \ll \min(n,m)\).

The optimal choice of rank and scaling parameters for LoRA methods are the subject of ongoing research. A proposal for the optimal choice of scaling factor  can be found, for example, in this publication \cite{kalajdzievski2023rank}. In \cite{Wang2022AdaMixMF}, a mixture of different PEFT methods was proposed to further enhancing performance.

\subsection{Federated Learning and Privacy}

Federated Learning was originally introduced by Google \cite{45648} as a decentralized training method for highly distributed (and private) data on mobile devices. However, it has quickly been adopted in other fields, particularly the healthcare sector \cite{engproc2023059230}, to make highly protected (patient) data from various sources available for collaborative training.

\begin{figure}
        \centering
        \includegraphics[width=5cm]{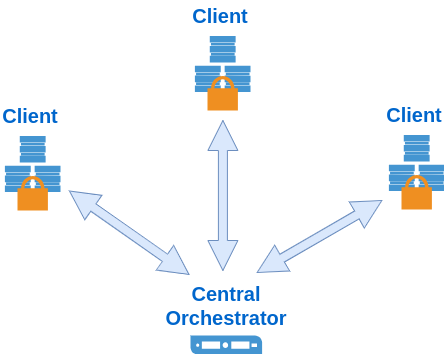}
        \caption{Federated Learning architecture with a central orchestrator
        and 3 clients with private datasets.}
        \label{fig:fl}
    \end{figure}

The method consists of two steps: a computation step, where clients locally train a model on their respective data, and an aggregation step, where the model parameters of the involved clients are aggregated on a central server. Thus, only the model parameters, not the data itself, leave the client.
There are also decentralized FL architectures that do not require a central server, but in this article, we focus exclusively on aggregation using a central orchestrator as illustrated in Fig. \ref{fig:fl}. 
A promising approach for federated PEFT methods in decentralized environments can be found in the following publication \cite{zhang-etal-2023-fedpetuning}.

The computation step and aggregation step together is called a communication round.
The distribution of the data can vary. In the simplest case, the data is independently and identically distributed (i.i.d.), but Federated Learning can also be successfully applied in the non-i.i.d. case \cite{https://doi.org/10.48550/arxiv.1806.00582}.
Especially in fine-tuning task federation can lead to a better generalization of the model and even be used for representation learning \cite{collins2022fedavgfinetuninglocal}.

\subsubsection{Privacy Enhancing Technologies} \label{pet}
Even if the data does not leave its location, it is possible that information could be revealed through the updates of the model parameters. To address this risk of disclosure, there are various privacy-enhancing technologies. In differential privacy \cite{10.1007/11787006_1, 10.1007/11681878_14}, noise is added to each parameter update, the amount of which is calculated based on a privacy budget and the sensitivity of the respective update function.

\section{Related Work - Existing Federation Approaches}
Federated Learning is applied in various scenarios. One example is a training scenario on mobile devices, where many devices contribute to the training, but not all are reliably available. Challenges like the high number of clients and their heterogeneity are addressed in these articles \cite{bai2024federatedfinetuninglargelanguage, cho2023heterogeneous}.  Another scenario is the healthcare \cite{doi:10.1021/acs.jcim.3c00799} or financial \cite{KAPSECKER2023101533} sector, where a smaller number of institutions agree on joint training on particularly protected data. 
In the following, we focus on the second scenario and assume a federated setup consisting of two clients, each with a private dataset, which is not accessible from outside the client. Both train the parameters $A_{1}, B_{1}$ and $A_{2}, B_{2}$ respectively. After each communication round, these parameters are aggregated and redistributed to the clients.

\subsection{Vanilla Federated Averaging}

The conventional Federated Averaging approach averages over the client adapters $A_{1}, A_{2}$ and $B_{1}, B_{2}$ by computing the (weighted) mean of the parameters. However, by definition, the LoRA parameters are a decomposition of the actual update parameter $\Delta W$. Averaging on each parameter separately introduces an error:

\begin{eqnarray}
    \tilde{A} = 0.5 \cdot (A_{1} + A_{2}) \nonumber\\
    \tilde{B} = 0.5 \cdot (B_{1} + B_{2}) \nonumber\\
    \tilde{B}\tilde{A} \neq 0.5 \cdot(B_{1}A_{1} + B_{2}A_{2}) \label{fedavg}
\end{eqnarray}

The precise aggregation would be the average of the products of the adaptors (right side of inequality \ref{fedavg}) and not the product of the averaged adaptors as in the vanilla approach (left side of inequality \ref{fedavg}). 
Despite from this additional introduced error we show below (in Eq. \ref{eq:DP_err}) that the vanilla approach is also not compatible with additive noise, which is a requirement for PET methods such as Differential Privacy.

\subsection{FFA-LoRA}
One way to overcome the above described error was proposed in Federated Freeze A LoRA (FFA-LoRA \cite{sun2024improving}). In this approach, only one of the adaptation matrices is trained, the other is frozen as well. This way, not only the aggregation error is eliminated, it turns out to be also compatible with differential privacy.

\paragraph{Differential Privacy}

Adding differential privacy to $A$ and $B$ leads to a quadratic noise term. To overcome this problem and the inequality of \eqref{fedavg}, the authors of \cite{sun2024improving} propose to freeze $A$ and train only $B$. This ensures that noise is added solely to B, thereby avoiding the non-linear effects that arise when two Gaussian noise terms are multiplied.
\begin{eqnarray}
    A_{\epsilon} = A + \eta_{\epsilon}\, &\eta_{\epsilon} \sim \mathcal{N}(0, \epsilon) \nonumber\\
    B_{\epsilon} = B + \eta_{\epsilon}\, &\eta_{\epsilon} \sim \mathcal{N}(0, \epsilon) \nonumber\\
    B_{\epsilon} \cdot A_{\epsilon} = B\cdot A + (A + B) \cdot \eta_{\epsilon} + \eta_{\epsilon}^2& \label{eq:DP_err}
\end{eqnarray}

We can see immediately that the error in equation \ref{eq:DP_err} is no longer normally distributed.
\begin{eqnarray}
B\cdot A + (A + B) \cdot \eta_{\epsilon} + \eta_{\epsilon}^2 \nsim  \mathcal{N}(0, \epsilon) \label{eq:DP_err_dist}
\end{eqnarray}

\section{Full Rank Aggregation (FRA) of LoRA Adapters}

The direct averaging of LoRA adapters introduces errors, as indicated in (\ref{fedavg}). We hypothesize that an alternative approach based on directly averaging the weight increments will reduce these errors. This method, termed Full Rank Aggregation LoRA (FRA-LoRA), is expected to improve the accuracy and effectiveness in the early training rounds of Federated Learning aggregation.

\begin{eqnarray*}
    \Delta W_{1} = B_{1}A_{1}\\
    \Delta W_{2} = B_{2}A_{2}
\end{eqnarray*}
\begin{eqnarray}
    \tilde{A} = 0.5 \cdot \texttt{concat}(A_{1}, A_{2}) \nonumber\\
    \tilde{B}= 0.5 \cdot \texttt{concat}(B_{1}, B_{2}) \nonumber\\
    \tilde{B}\tilde{A} = 0.5 \cdot  (\Delta W_{1}+\Delta W_{2})=\Delta W \label{concatavg}
\end{eqnarray}
This approach is also compatible with Differential Privacy as the noise can be added directly on the high-dimensional aggregation (eq. \ref{concatavg}), such that non-linear effects that come with the product of the low rank adapters are avoided.

In a setup with a small number of clients the concatenation of low rank matrices has still a rather low rank and thus the matrix multiplication stays 
$ \mathcal{O} ( r \times (n+m) )$, 
where $r$ is the rank and $m \times r$ is the dimensionality of $B$ and $r \times n$ the dimensionality of $A$.
However, the return value of the federated aggregation must have the same dimensionality as the unaggregated contributors, which is obviously not the case for the concatenations. Thus, we need an additional step to reduce the dimensionality of aggregated adapters back to the original. 
The main requirement we want to achieve with the aggregation of the federated parameters is a close approximation of the aggregated weight increment $\Delta W$. This is the product of the concatenated matrices (\ref{concatavg}) but can be as well decomposed again in low rank matrices of rank $r$ like the original federated LoRA adapters.
For this purpose we perform a singular value decomposition (SVD) of $\Delta W$ and keep only the eigenvectors that belong to the $r$ highest singular values.

The singular value decomposition of the weight increment has the following structure:
\begin{eqnarray}
    \texttt{SVD}\left(\Delta W\right) &=& UDV \nonumber\\
    &=&  \sum_{j= 0}^R u_{ij} d_j v_{jk},\label{svd}
\end{eqnarray}
where R is the rank of the original matrix $\Delta W$.
For the low rank decomposition, we keep only the first $r \leq R $ singular values (in order of there absolute values) and obtain

\begin{eqnarray}
    LoRA(\Delta W) = \sum_{j= 0}^r u_{ij} d_j v_{jk} = BA \label{dimred}
\end{eqnarray}

A similar aggregation approach can be found in \cite{babakniya2023slora}; however the described algorithm consists of two stages first, a full fine-tuning is performed on a precomputed mask, and in the second step, LoRA is applied. The aforementioned dimensionality reduction using SVD (eq. \ref{svd}-\ref{dimred}) is only used between the two stages.
In FRA-LoRA, full fine-tuning of selected weight matrices does not take place; instead, LoRA is applied to all model layers, but the aggregation is performed on the full matrices and then SVD is applied before propagating the low rank adaptors back to the clients.

\paragraph{Differential Privacy}
If Differential Privacy is applied, noise is added to the aggregated result. With FRA LoRA we are able to add the noise directly to the aggregated gradient before decomposing it to A and B. So when the results are propagated back to the clients and multiplied there, we get back the normal (or Laplace) distributed noise that was originally added.

\begin{eqnarray}
    LoRA(\Delta W + \eta_{\epsilon})
     = BA + \eta_{\epsilon}, \label{dimred_noise}\\
     \eta_{\epsilon} \sim \mathcal{N}(0, \epsilon)   \nonumber
\end{eqnarray}

\section{Experiments}

In the following, we examine the presented aggregation methods using the General Language Understanding Evaluation (GLUE) benchmark datasets \cite{wang2019glue} as examples. These are common benchmarks for LM tasks.
The datasets selected for this article were:
\begin{enumerate}
    \item Multi-Genre Natural Language Inference (MNLI) \cite{williams-etal-2018-broad}: this dataset contains a collection of sentence pairs (a premise and a hypothesis) and a label that indicates whether the hypothesis is entailed by the premise, contradicted or neutral. The training dataset has 392702 data points. For MNLI are two validation sets available, we chose the validation-matched, which was generated from the same source as the training set and contains 9815 data points.
    \item Stanford Sentiment Treebank (SST2) \cite{conf/emnlp/SocherPWCMNP13}: This dataset contains  movie reviews that are labeled with human annotations of their sentiment. The training dataset has 67349 data points, the validation set 872.
\end{enumerate}
As performance metrics was used in both examples the accuracy metric according to the guidelines of the GLUE tasks. 

For all examples, we used the RoBERTa-base \cite{DBLP:journals/corr/abs-1907-11692} checkpoint from HuggingFace as pre-trained model, which is with 12 4647 170 parameters relatively small. With LoRA the trainable parameters even reduce to 1 774 084.
We take the centralised training performace of RoBERTa-base as a baseline, and compare our method with related works from the literature.

We first show results against the ideal scenario of equally split datasets of i.i.d. data. We then extend our experiments to assess the impact of heterogeneous dataset splits. In this more realistic scenario, we partition the data such that target classes and data sizes differ across sites.
Without loss of generality the aggregation methods are also applicable to a large number of federated instances.

\subsection{Federated Training on balanced Datasets}
First, we test all three methods in a federated setting by dividing the datasets into two arbitrary i.i.d. sets to simulate federated training. The parameters are aggregated after each epoch following the FedAvg training regime. The hyperparameters for the experiments below were chosen as follows: weight decay 0.005, learning rate 1e-4, and dropout 0.2. The training was conducted over 30 communication rounds.
For MNLI we specified a maximum number of steps per epoch of 5000, for the SST2 dataset, which is smaller, we trained over the entire datasets in each epoch.

    \begin{figure}
        \centering
        \includegraphics[width=5cm]{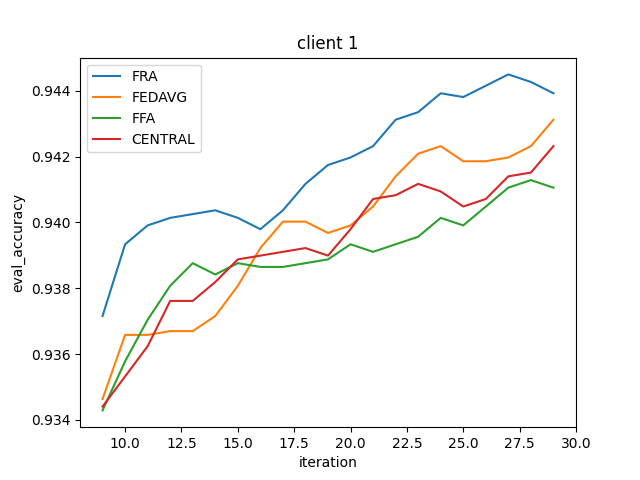}
        \includegraphics[width=5cm]{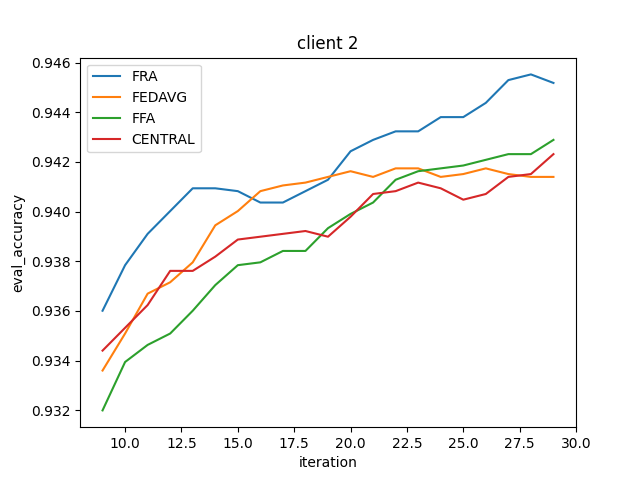}
        \caption{SST2 Dataset (i.i.d. split with balanced classes): Evaluation set accuracy on each client per iteration over a rolling average of 7.}
        \label{fig:sst2-balanced}
    \end{figure}

    \begin{table}
        \centering
        \begin{tabular}{|l||cccc|}
            \hline
            &FRA & FEDAVG  & FFA & CENTRALISED\\   
            \hline
            \hline
            SST2 & 94.9541  & 94.8394 & 94.7248 &  94.7248\\
            MNLI & 85.3795 &  85.2471  &84.1773 &86.0316\\
     \hline
        \end{tabular}
        \caption{Evaluation set accuracy of best model.}
        \label{tab:balanced}
    \end{table}
    \begin{figure}
        \centering
        \includegraphics[width=5cm]{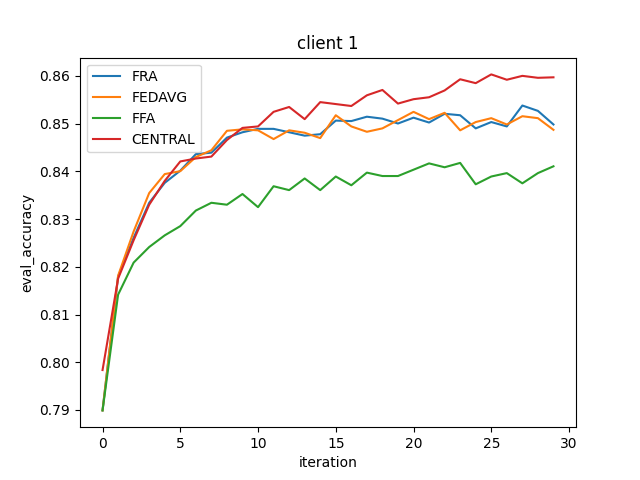}
        \includegraphics[width=5cm]{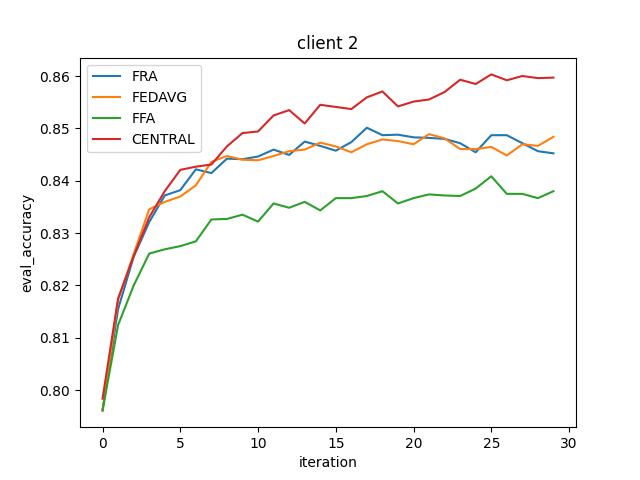}
        \caption{MNLI Dataset (i.i.d. split with balanced classes): Evaluation set accuracy on each client per iteration.}
        \label{fig:mnli-balanced}
    \end{figure}
As shown in Fig. (\ref{fig:sst2-balanced}), the evaluation accuracy improves steadily over the first 15-20 federation rounds, but then we see the accuracy drop as the model starts to overfit. The accuracy values of the SST2 example are closer to each other, such that we chose to plot it with a rolling average of 7 for a better visibility.
For the larger MNLI dataset (Fig. \ref{fig:mnli-balanced}) FRA-LoRA and vanilla Federated Averaging are both close to the centralised results (c.f. Tab. \ref{tab:balanced}). However, as shown in Eq. \ref{eq:DP_err_dist}
FedAvg can not be applied in a privacy preserving way and a centralised computation is anyhow only feasible, if data can be freely accessed. That leaves only FFA and FRA-LoRA as options for a privacy preserving training.

\subsection{Federated Training on imbalanced Datasets}

Experiments on imbalanced datasets further confirm the findings from the i.i.d. setting. The results in terms of accuracy are listed in table \ref{tab:imbalanced}. The datasets were distributed according to their labels as given in table \ref{tab:sst2-dist}.

\paragraph{SST2}
    \begin{table}
        \centering
        \begin{tabular}{|l||cc|}
            \hline
            &client 1& client 2 \\   
            \hline
            \hline
            Label 0 & 0.9 & 0.1 \\
            Label 1 & 0.1 & 0.9\\
            \hline
        \end{tabular}
        \caption{Label Distribution over Clients.}
        \label{tab:sst2-dist}
    \end{table}

    \begin{figure}
        \centering
        \includegraphics[width=5cm]{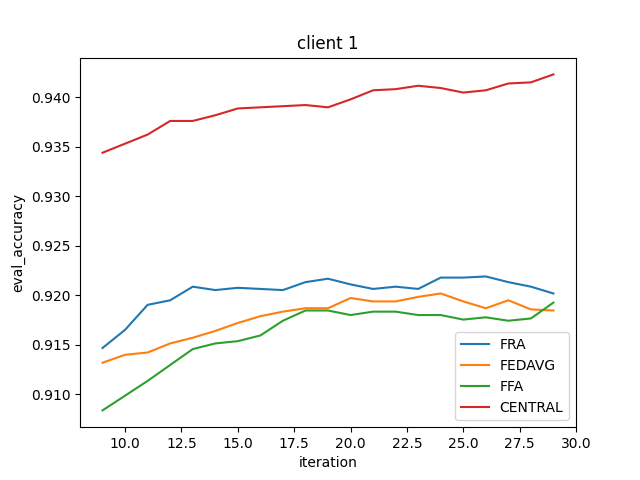}
        \includegraphics[width=5cm]{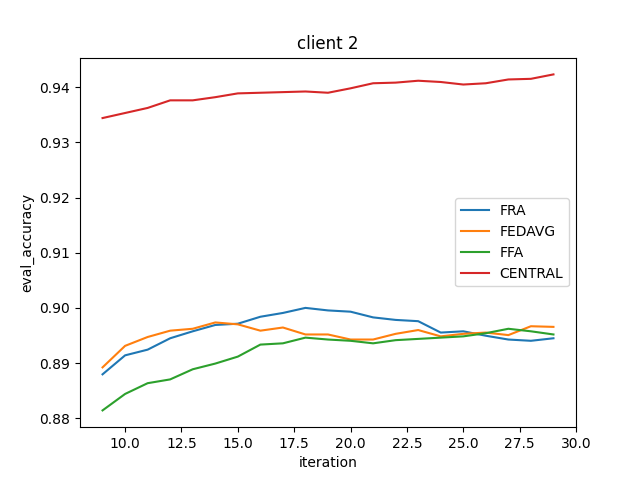}
        \caption{SST2 Dataset (split with imbalanced classes): Evaluation set accuracy on each client per iteration over a rolling average of 7.}
        \label{fig:sst2-imbalanced}
    \end{figure}

    \begin{table}
        \centering
        \begin{tabular}{|l||cccc|}
            \hline
            &FRA & FEDAVG  & FFA & CENTRALISED\\   
            \hline
            \hline
            SST2 & 92.7752  & 93.1193 &  92.5459 & 94.7248\\
            MNLI & 85.5018, & 85.7769& 83.5354 & 86.0316\\
            \hline
        \end{tabular}
        \caption{Evaluation set accuracy of best model.}
        \label{tab:imbalanced}
    \end{table}
    The evolution of evaluation set accuracy over the communication rounds (Fig. \ref{fig:sst2-imbalanced}) show, that FRA-LoRA is performing well in the first 15-20 rounds but then accuracy drops and the model starts to overfit. In all experiments, we used a dropout factor of 0.2, weight decay 0.005 and a learning rate of 0.0001.
    For the smaller dataset (SST2) we observed that FRA-LoRA and vanilla Federated Averaging are improving the evaluation set accuracy fast in the first iterations but start to overfit at some point, whereas FFA-LoRA shows improvements in the later iterations. 
    That FRA-LoRA starts overfitting earlier is probably due to the double size of parameters compared to FFA. Also recent research \cite{pmlr-v235-hayou24a} found out that the A matrix contributes less for smaller learning rates, i.e. in later iteration.
    In this case, a combination of FRA-LoRA and FFA would be the method of choice. 
    \paragraph{MNLI}

    The imbalanced distribution of MNLI dataset is composed according to the following distribution as given in table \ref{tab:mnli-dist}.
        \begin{table}
        \centering
        \begin{tabular}{|l||cc|}
            \hline
            &client 1& client 2 \\   
            \hline
            \hline
            Label 0 & 0.7 & 0.3 \\
            Label 1 & 0.2 & 0.8\\
            Label 2 & 0.2 & 0.8\\
            \hline
        \end{tabular}
        \caption{Label Distribution over Clients.}
        \label{tab:mnli-dist}
    \end{table}

    \begin{figure}
        \centering
        \includegraphics[width=5cm]{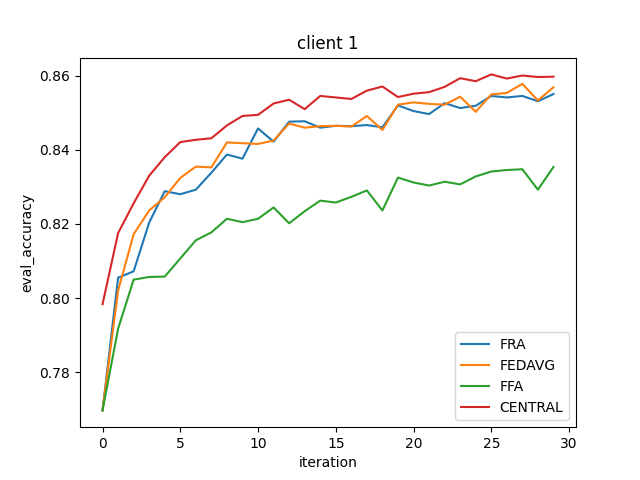}
        \includegraphics[width=5cm]{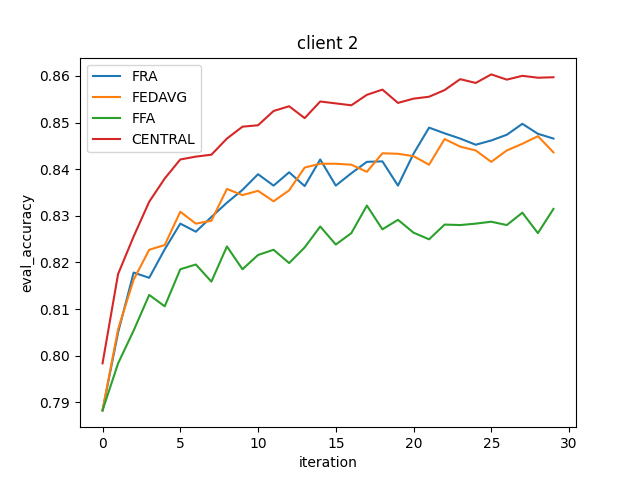}
        \caption{MNLI Dataset (split with imbalanced classes): Evaluation set accuracy on each client per iteration.}
        \label{fig:mnli-imbalanced}
    \end{figure}

    In the MNLI example, we find that vanilla Federated Averaging and FRA-LoRA perform equally well, while FFA is clearly outperformed, results are shown in Fig. \ref{fig:mnli-imbalanced}. In use cases with rather lenient data protection requirements FedAvg and FRA-LoRA can be considered equal. However, only FRA-LoRA is compatible with additional noise. 
    \paragraph{Error Analysis}
    All approaches contain a deviation from the exact gradient in the aggregation which is introduced by the low dimesional projection of $A$ and $B$ introduced by LoRA. 
    In case of FedAvg we get an additional error that arises from inequality (\ref{fedavg}), 
    
        \begin{eqnarray*}
            err_{\texttt{FedAvg}} = 0.25 \cdot (A_{1}-A_{2})\cdot(B_{2}-B_{1})
        \end{eqnarray*}

    while in FRA-LoRA, the error is again due to the approximation:
    
        \begin{eqnarray*}
            LoRA(\Delta W) = \sum_{j= 0}^r u_{ij} d_j v_{jk} = BA \\
            \neq  \sum_{j= 0}^R u_{ij} d_j v_{jk} = \Delta W \\
             err_{\texttt{FRA-LoRA}} = \sum_{j= r+1}^R u_{ij} d_j v_{jk}
        \end{eqnarray*}

\begin{figure}
        \centering
        \includegraphics[width=5cm]{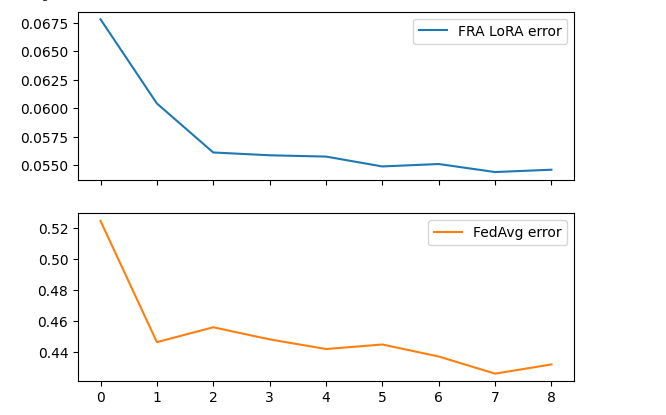}
        \caption{Errors $err_{\texttt{FRA-LoRA}}$ and $err_{\texttt{FedAvg}}$   for the first 10 iterations of the MNLI imbalanced example. FRA-LoRA errors are an order of magnitude lower in absolute terms. 
        }
        \label{fig:errors}
    \end{figure}

    This error is introduced by the rank reduction and is in the same order of magnitude as the error that is already introduced by the LoRA method itself.
    A comparison of both errors for the first 10 iterations of the MNLI imbalanced experiment is shown in Fig. \ref{fig:errors}.
    However, both errors in this case seem reasonable small such that the training performance is not negatively affected.

\section{Conclusion and Outlook}
Based on the preceding experiments, we believe that FRA-LoRA can contribute to the federated fine-tuning of LLMs, especially under the following conditions:

\begin{enumerate}
    
\item The number of clients is small and the rank of the adaptors was chosen 
significantly smaller than the dimension of the original weight matrix,

\item  If the fine-tuning data is large and the risk of overfitting is small, it can speed up the training by reaching a higher accuracy level faster.
FRA-LoRA achieves a good accuracy level faster, but in some cases, we observed that it is more prone to overfitting compared to FFA-LoRA. This is easily understandable since FRA-LoRA trains twice the number of parameters. 
Preliminary investigations suggest that the overfitting tendencies can be mitigated by adjustments to the learning rate scheduler, weight decay as well as improved hyper-parameter search methods. 
In future work, it would certainly be worthwhile to explore a combination of both methods. 
Use FRA in the early rounds where the model is less stable and accuracy of gradients is more important, then simplify to FFA once the training is stable and the A matrix is sufficiently trained so as not to impact training performance.
These findings align with recent work on LoRA convergence \cite{pmlr-v235-hao24a}, showing that as the training stabilises the A matrix has lower impact on training. Therefore it naturally follows that using FLA-LoRA should improve model convergence in the early stages of training. 

\item  Fine-tuning data is sensitive and privacy enhancing methods are required.
Since in Federated Learning the training data is often sensitive, privacy preserving training methods are of high importance and the compatibility of FRA-LoRA with Differential Privacy is a clear advantage over the vanilla Federated Averaging.

\item A very recent promising improvement of LoRA itself is proposed here \cite{yan2024federa}. The authors propose to initialize A and B in the training rounds
by an SVD of the weight update. This proceeding fits very well with our approach of aggregation.
\end{enumerate}

Our aggregation method is currently based on Federated Averaging. In future work, it should be investigated how the approach performs with other aggregation techniques, such as Federated Optimization \cite{li2020federated}, Scaffolding \cite{DBLP:journals/corr/abs-1910-06378} and  variants of Federated Averaging, as used, for example, in \cite{jiang2023lowparameter}.

\section*{Acknowledgment}
We would like to thank everyone who contributed to the success of this article. In particular, our colleague Jan Stücke for his valuable feedback and proof-reading. Special thanks go to the entire Apheris team for the inspiring atmosphere and exchange on all federation related topics.
\bibliographystyle{plain}
\bibliography{bibtex}

\begin{thebibliography}{10}

\bibitem{babakniya2023slora}
Sara Babakniya, Ahmed~Roushdy Elkordy, Yahya~H. Ezzeldin, Qingfeng Liu,
  Kee-Bong Song, Mostafa El-Khamy, and Salman Avestimehr.
\newblock Slora: Federated parameter efficient fine-tuning of language models,
  2023.

\bibitem{bai2024federatedfinetuninglargelanguage}
Jiamu Bai, Daoyuan Chen, Bingchen Qian, Liuyi Yao, and Yaliang Li.
\newblock Federated fine-tuning of large language models under heterogeneous
  tasks and client resources, 2024.

\bibitem{cho2023heterogeneous}
Yae~Jee Cho, Luyang Liu, Zheng Xu, Aldi Fahrezi, Matt Barnes, and Gauri Joshi.
\newblock Heterogeneous lo{RA} for federated fine-tuning of on-device
  foundation models.
\newblock In {\em International Workshop on Federated Learning in the Age of
  Foundation Models in Conjunction with NeurIPS 2023}, 2023.

\bibitem{collins2022fedavgfinetuninglocal}
Liam Collins, Hamed Hassani, Aryan Mokhtari, and Sanjay Shakkottai.
\newblock Fedavg with fine tuning: Local updates lead to representation
  learning, 2022.

\bibitem{engproc2023059230}
Pallavi Dhade and Prajakta Shirke.
\newblock Federated learning for healthcare: A comprehensive review.
\newblock {\em Engineering Proceedings}, 59(1), 2023.

\bibitem{10.1007/11787006_1}
Cynthia Dwork.
\newblock Differential privacy.
\newblock In Michele Bugliesi, Bart Preneel, Vladimiro Sassone, and Ingo
  Wegener, editors, {\em Automata, Languages and Programming}, pages 1--12,
  Berlin, Heidelberg, 2006. Springer Berlin Heidelberg.

\bibitem{10.1007/11681878_14}
Cynthia Dwork, Frank McSherry, Kobbi Nissim, and Adam Smith.
\newblock Calibrating noise to sensitivity in private data analysis.
\newblock In Shai Halevi and Tal Rabin, editors, {\em Theory of Cryptography},
  pages 265--284, Berlin, Heidelberg, 2006. Springer Berlin Heidelberg.

\bibitem{han2024parameterefficientfinetuninglargemodels}
Zeyu Han, Chao Gao, Jinyang Liu, Jeff Zhang, and Sai~Qian Zhang.
\newblock Parameter-efficient fine-tuning for large models: A comprehensive
  survey, 2024.

\bibitem{pmlr-v235-hao24a}
Yongchang Hao, Yanshuai Cao, and Lili Mou.
\newblock Flora: Low-rank adapters are secretly gradient compressors.
\newblock In Ruslan Salakhutdinov, Zico Kolter, Katherine Heller, Adrian
  Weller, Nuria Oliver, Jonathan Scarlett, and Felix Berkenkamp, editors, {\em
  Proceedings of the 41st International Conference on Machine Learning}, volume
  235 of {\em Proceedings of Machine Learning Research}, pages 17554--17571.
  PMLR, 21--27 Jul 2024.

\bibitem{pmlr-v235-hayou24a}
Soufiane Hayou, Nikhil Ghosh, and Bin Yu.
\newblock {L}o{RA}+: Efficient low rank adaptation of large models.
\newblock In Ruslan Salakhutdinov, Zico Kolter, Katherine Heller, Adrian
  Weller, Nuria Oliver, Jonathan Scarlett, and Felix Berkenkamp, editors, {\em
  Proceedings of the 41st International Conference on Machine Learning}, volume
  235 of {\em Proceedings of Machine Learning Research}, pages 17783--17806.
  PMLR, 21--27 Jul 2024.

\bibitem{doi:10.1021/acs.jcim.3c00799}
Wouter Heyndrickx, Lewis Mervin, Tobias Morawietz, Noé Sturm, Lukas Friedrich,
  Adam Zalewski, Anastasia Pentina, Lina Humbeck, Martijn Oldenhof, Ritsuya
  Niwayama, Peter Schmidtke, Nikolas Fechner, Jaak Simm, Adam Arany, Nicolas
  Drizard, Rama Jabal, Arina Afanasyeva, Regis Loeb, Shlok Verma, Simon
  Harnqvist, Matthew Holmes, Balazs Pejo, Maria Telenczuk, Nicholas Holway,
  Arne Dieckmann, Nicola Rieke, Friederike Zumsande, Djork-Arné Clevert,
  Michael Krug, Christopher Luscombe, Darren Green, Peter Ertl, Peter Antal,
  David Marcus, Nicolas Do~Huu, Hideyoshi Fuji, Stephen Pickett, Gergely Acs,
  Eric Boniface, Bernd Beck, Yax Sun, Arnaud Gohier, Friedrich Rippmann, Ola
  Engkvist, Andreas~H. Göller, Yves Moreau, Mathieu~N. Galtier, Ansgar
  Schuffenhauer, and Hugo Ceulemans.
\newblock Melloddy: Cross-pharma federated learning at unprecedented scale
  unlocks benefits in qsar without compromising proprietary information.
\newblock {\em Journal of Chemical Information and Modeling}, 64(7):2331--2344,
  2024.
\newblock PMID: 37642660.

\bibitem{houlsby2019parameterefficient}
Neil Houlsby, Andrei Giurgiu, Stanislaw Jastrzebski, Bruna Morrone, Quentin
  de~Laroussilhe, Andrea Gesmundo, Mona Attariyan, and Sylvain Gelly.
\newblock Parameter-efficient transfer learning for nlp, 2019.

\bibitem{hu2021lora}
Edward~J. Hu, Yelong Shen, Phillip Wallis, Zeyuan Allen-Zhu, Yuanzhi Li, Shean
  Wang, Lu~Wang, and Weizhu Chen.
\newblock Lora: Low-rank adaptation of large language models, 2021.

\bibitem{jiang2023lowparameter}
Jingang Jiang, Xiangyang Liu, and Chenyou Fan.
\newblock Low-parameter federated learning with large language models, 2023.

\bibitem{kalajdzievski2023rank}
Damjan Kalajdzievski.
\newblock A rank stabilization scaling factor for fine-tuning with lora, 2023.

\bibitem{KAPSECKER2023101533}
Maximilian Kapsecker, Daniel~N. Nugraha, Christoph Weinhuber, Nicholas Lane,
  and Stephan~M. Jonas.
\newblock Federated learning with swift: An extension of flower and performance
  evaluation.
\newblock {\em SoftwareX}, 24:101533, 2023.

\bibitem{DBLP:journals/corr/abs-1910-06378}
Sai~Praneeth Karimireddy, Satyen Kale, Mehryar Mohri, Sashank~J. Reddi,
  Sebastian~U. Stich, and Ananda~Theertha Suresh.
\newblock {SCAFFOLD:} stochastic controlled averaging for on-device federated
  learning.
\newblock {\em CoRR}, abs/1910.06378, 2019.

\bibitem{45648}
Jakub Konečný, H.~Brendan McMahan, Felix~X. Yu, Peter Richtarik,
  Ananda~Theertha Suresh, and Dave Bacon.
\newblock Federated learning: Strategies for improving communication
  efficiency.
\newblock In {\em NIPS Workshop on Private Multi-Party Machine Learning}, 2016.

\bibitem{li2020federated}
Tian Li, Anit~Kumar Sahu, Manzil Zaheer, Maziar Sanjabi, Ameet Talwalkar, and
  Virginia Smith.
\newblock Federated optimization in heterogeneous networks, 2020.

\bibitem{DBLP:journals/corr/abs-1907-11692}
Yinhan Liu, Myle Ott, Naman Goyal, Jingfei Du, Mandar Joshi, Danqi Chen, Omer
  Levy, Mike Lewis, Luke Zettlemoyer, and Veselin Stoyanov.
\newblock Roberta: {A} robustly optimized {BERT} pretraining approach.
\newblock {\em CoRR}, abs/1907.11692, 2019.

\bibitem{mcmahan2023communicationefficient}
H.~Brendan McMahan, Eider Moore, Daniel Ramage, Seth Hampson, and
  Blaise~Agüera y~Arcas.
\newblock Communication-efficient learning of deep networks from decentralized
  data, 2023.

\bibitem{pfeiffer-etal-2020-adapterhub}
Jonas Pfeiffer, Andreas R{\"u}ckl{\'e}, Clifton Poth, Aishwarya Kamath, Ivan
  Vuli{\'c}, Sebastian Ruder, Kyunghyun Cho, and Iryna Gurevych.
\newblock {A}dapter{H}ub: A framework for adapting transformers.
\newblock In Qun Liu and David Schlangen, editors, {\em Proceedings of the 2020
  Conference on Empirical Methods in Natural Language Processing: System
  Demonstrations}, pages 46--54, Online, October 2020. Association for
  Computational Linguistics.

\bibitem{conf/emnlp/SocherPWCMNP13}
Richard Socher, Alex Perelygin, Jean Wu, Jason Chuang, Christopher~D. Manning,
  Andrew~Y. Ng, and Christopher Potts.
\newblock Recursive deep models for semantic compositionality over a sentiment
  treebank.
\newblock In {\em EMNLP}, pages 1631--1642. ACL, 2013.

\bibitem{sun2024improving}
Youbang Sun, Zitao Li, Yaliang Li, and Bolin Ding.
\newblock Improving lo{RA} in privacy-preserving federated learning.
\newblock In {\em The Twelfth International Conference on Learning
  Representations}, 2024.

\bibitem{wang2019glue}
Alex Wang, Amanpreet Singh, Julian Michael, Felix Hill, Omer Levy, and
  Samuel~R. Bowman.
\newblock {GLUE}: A multi-task benchmark and analysis platform for natural
  language understanding.
\newblock 2019.
\newblock In the Proceedings of ICLR.

\bibitem{Wang2022AdaMixMF}
Yaqing Wang, Subhabrata Mukherjee, Xiaodong Liu, Jing Gao, and Jianfeng Gao.
\newblock Adamix: Mixture-of-adaptations for parameter-efficient model tuning.
\newblock In {\em Conference on Empirical Methods in Natural Language
  Processing}, 2022.

\bibitem{williams-etal-2018-broad}
Adina Williams, Nikita Nangia, and Samuel Bowman.
\newblock A broad-coverage challenge corpus for sentence understanding through
  inference.
\newblock In Marilyn Walker, Heng Ji, and Amanda Stent, editors, {\em
  Proceedings of the 2018 Conference of the North {A}merican Chapter of the
  Association for Computational Linguistics: Human Language Technologies,
  Volume 1 (Long Papers)}, pages 1112--1122, New Orleans, Louisiana, June 2018.
  Association for Computational Linguistics.

\bibitem{yan2024federa}
Yuxuan Yan, Shunpu Tang, Zhiguo Shi, and Qianqian Yang.
\newblock Federa: Efficient fine-tuning of language models in federated
  learning leveraging weight decomposition.
\newblock {\em arXiv preprint arXiv:2404.18848}, 2024.

\bibitem{zhang-etal-2023-fedpetuning}
Zhuo Zhang, Yuanhang Yang, Yong Dai, Qifan Wang, Yue Yu, Lizhen Qu, and Zenglin
  Xu.
\newblock {F}ed{PET}uning: When federated learning meets the
  parameter-efficient tuning methods of pre-trained language models.
\newblock In Anna Rogers, Jordan Boyd-Graber, and Naoaki Okazaki, editors, {\em
  Findings of the Association for Computational Linguistics: ACL 2023}, pages
  9963--9977, Toronto, Canada, July 2023. Association for Computational
  Linguistics.

\bibitem{https://doi.org/10.48550/arxiv.1806.00582}
Yue Zhao, Meng Li, Liangzhen Lai, Naveen Suda, Damon Civin, and Vikas Chandra.
\newblock Federated learning with non-iid data.
\newblock 2018.

\end{thebibliography}

\end{document}